\documentclass[opre,nonblindrev]{informs4_hide}

\OneAndAHalfSpacedXI 

\usepackage{amsmath,amsfonts,amssymb}
\usepackage{mathtools}
    \allowdisplaybreaks
\usepackage{bm}
\usepackage{bbm}
\usepackage[mathscr]{euscript}
\usepackage{enumitem}
\usepackage{xcolor}
\usepackage{graphicx}
\usepackage{multirow}
\usepackage{booktabs}
\usepackage{microtype}
\usepackage{fix-cm}
\usepackage[ruled,vlined]{algorithm2e}
\usepackage{tikz}
\usepackage{smile}
\usepackage{subcaption}

\DeclareMathOperator{\tr}{tr}

\DeclareMathOperator{\rank}{rank}

\newcommand{\one}{\mathbbm{1}}

\usepackage{natbib}
 \bibpunct[, ]{(}{)}{,}{a}{}{,}%
 \def\bibfont{\small}%
\TheoremsNumberedThrough     
\ECRepeatTheorems

\EquationsNumberedThrough    

\begin{document}


\RUNAUTHOR{Wang and Zhang}

\RUNTITLE{On the Suboptimality of GP-UCB}

\TITLE{On the Suboptimality of GP-UCB under Polynomial Effective Optimism}

\ARTICLEAUTHORS{
\AUTHOR{Wenjia Wang}
\AFF{Department of Industrial Systems Engineering and Management, National University of Singapore, Singapore 117576, \EMAIL{wenjiawang@nus.edu.sg}}
\AUTHOR{Xiaowei Zhang}
\AFF{Department of Industrial Engineering and Decision Analytics, The Hong Kong University of Science and Technology, Clear Water Bay, Hong Kong SAR, \EMAIL{xiaoweiz@ust.hk}}
}

\ABSTRACT{Gaussian process upper confidence bound (GP-UCB) is widely used for sequential optimization of expensive black-box functions. Although many upper bounds on its cumulative regret have been established in the literature, whether GP-UCB is minimax optimal remains open. 
We study this question through the effective optimism level, defined as the product of the exploration coefficient and the regularization parameter in kernel ridge regression. Under a uniform confidence assumption, we prove a new regret lower bound for GP-UCB with Mat\'ern kernels. The bound shows that polynomial growth of the effective optimism level, up to logarithmic factors, rules out the minimax-optimal regret rate. Since this is the regime covered by most existing analyses, our result identifies a concrete obstacle to proving minimax optimality for standard GP-UCB. More broadly, it suggests that the gap between current upper bounds and minimax lower bounds may reflect a real limitation of the algorithm, not only of the analysis. 
}



\maketitle

\section{Introduction}\label{sec:introduction}

Optimizing expensive black-box functions from noisy evaluations is a central problem in many areas of science and engineering. Examples include simulation optimization \citep{PearcePoloczekBranke22}, online experimentation \citep{LethamBakshy19}, engineering design \citep{Frazier2016}, and hyperparameter tuning in machine learning \citep{FranceschiDoniniPerroneKleinArchambeauSeegerPontilFrasconi25}; see also \citet{ShahriariSwerskyWangAdamsdeFreitas16} for an overview. In these settings, each evaluation is often costly, so the goal is to identify near-optimal points using as few samples as possible. Gaussian process upper confidence bound (GP-UCB) is one of the most widely used Bayesian optimization methods for this task \citep{Frazier18}. It places a Gaussian process prior on the objective function and, at each round, selects the point that maximizes the posterior mean plus an exploration bonus proportional to the posterior standard deviation. 
This rule is simple to implement, easy to interpret, and naturally balances exploration and exploitation.

The theoretical performance of GP-UCB is typically measured by cumulative regret. Since the seminal work of \citet{srinivas2010gaussian}, a large literature has established sublinear regret guarantees under smoothness assumptions on the objective \citep{chowdhury2017kernelized,JanzBurtGonzalez20,VakiliKhezeliPicheny21,WhitehouseWuRamdas23}. For functions in the reproducing kernel Hilbert space (RKHS) associated with Mat\'ern kernels, these results have substantially narrowed the gap between available upper bounds for GP-UCB and the minimax lower bounds for optimization over the same classes. This naturally raises the question of whether GP-UCB can achieve minimax-optimal regret, at least up to logarithmic factors.

This paper studies a barrier to such guarantees. Our focus is on two quantities that appear throughout existing analyses: the exploration coefficient \(\beta_t\), which scales the posterior variance in the acquisition rule, and the regularization parameter \(\rho_t\) in kernel ridge regression. Their product,
\[
L_t \coloneqq \rho_t \beta_t,
\]
captures the algorithm's overall level of optimism under uncertainty. We refer to \(L_t\) as the \emph{effective optimism level}. In existing analyses of GP-UCB, it typically grows polynomially in \(t\), up to logarithmic factors.

Our main result shows that this growth can impose a fundamental limitation. We establish a regret lower bound for GP-UCB over RKHSs associated with Mat\'ern kernels, showing that polynomial effective optimism rules out minimax optimality. In particular, for Mat\'ern kernels with smoothness parameter \(\nu\), there exist hard instances for which GP-UCB incurs regret of order
$
T^{\frac{\nu+d}{2\nu+d}} L_T^{\frac{\nu}{2\nu+d}},
$
where \(d\) denotes the dimension of the domain. 
Thus, if \(L_T\) grows polynomially in \(T\), up to logarithmic factors, then GP-UCB cannot attain the minimax-optimal rate \(T^{\frac{\nu+d}{2\nu+d}}\) for Mat\'ern RKHSs \citep{scarlett2017lower}.

Our result offers a new perspective on the current state of GP-UCB theory. Recent work has largely sought to tighten existing analyses so as to match minimax lower bounds. Our lower bound indicates that when the effective optimism level remains polynomial, the residual gap need not be a technical artifact of the analysis. It may instead reflect a genuine limitation of the sampling behavior induced by standard choices of exploration and regularization parameters.

The rest of the paper is organized as follows. Section~\ref{sec:problem} introduces the model and assumptions. Section~\ref{sec:mainresult} states the regret lower bound. Section~\ref{sec:construction} constructs the hard instances for Mat\'ern RKHSs. Section~\ref{sec:analysis} proves the lower bound. Section~\ref{sec:conclusion} concludes.

\section{Problem Formulation}\label{sec:problem}

We study the problem of maximizing a function \(f\) over a feasible region \(\cD\subset\RR^d\) through adaptive sampling. 
Without loss of generality, we take \(\cD=[0,1]^d\). 
At each round \(t\), an algorithm \(\cA\) selects a point \(\bx_t\in\cD\) based on past observations and then receives a noisy sample 
  $y_t=f(\bx_t)+\varepsilon_t$.
We measure the performance of $\cA$ by its cumulative regret, defined as
\[
  \cR(T;f, \mathcal{A}) \coloneqq \sum_{t=1}^T\left[f(\bx^\star) - f(\bx_t)\right],
\]
where $T$ is the sampling budget and $\bx^\star\in\argmax_{\bx\in \cD} f(\bx)$ is a maximizer of $f$. 

\subsection{GP-UCB}

GP-UCB, introduced by \citet{srinivas2010gaussian}, is widely used in practice because of its simple form and strong empirical performance. The algorithm is most naturally described from a Bayesian perspective under Gaussian noise. Our analysis, however, is entirely frequentist and allows for more general noise assumptions. In particular, minimax optimality is understood throughout in the frequentist sense, not relative to any prior distribution on \(f\).

We now recall the Bayesian construction used to define the algorithm. 
Suppose \(f\) is equipped with a zero-mean Gaussian process prior with covariance function \(k:\cD\times\cD\to\RR\), also called the kernel. That is, for any finite set \(\{\bx_1,\ldots,\bx_n\}\subset\cD\), the random vector \((f(\bx_1),\ldots,f(\bx_n))\) is multivariate normal with mean zero and covariance matrix \((k(\bx_i,\bx_j))_{1\le i,j\le n}\).

Assume for the moment that the observation noise variables \(\varepsilon_t\) are independent and identically distributed (i.i.d.) normal with mean $0$ and \emph{known} variance \(\sigma^2\).
Given observations \(\{(\bx_1,y_1),\ldots,(\bx_t,y_t)\}\), the posterior distribution of \(f\) is again a Gaussian process, with mean function \(\mu_t\) and covariance function \(k_t\) given by
\begin{align}
\mu_t(\bx)
=&\,
\kb_t^\intercal(\bx)(\Kb_t+\sigma^2 \Ib_t)^{-1}\by_t,
\label{eq:posterior-mean}
\\
k_t(\bx,\bx')
=&\,
k(\bx,\bx')-\kb_t^\intercal(\bx)(\Kb_t+\sigma^2 \Ib_t)^{-1}\kb_t(\bx'),
\label{eq:posterior-cov}
\end{align}
where \(\kb_t(\bx)=(k(\bx,\bx_1),\ldots,k(\bx,\bx_t))^\intercal\), \(\Kb_t=(k(\bx_i,\bx_j))_{i,j\le t}\), and \(\by_t=(y_1,\ldots,y_t)^\intercal\).

Motivated by the upper confidence bound principle in multi-armed bandits \citep{AuerCesa-BianchiFischer02}, GP-UCB selects the next sampling point \(\bx_{t+1}\) according to
\begin{equation*}
\bx_{t+1}
=
\argmax_{\bx\in\cD}
\left\{
\mu_t(\bx)+\beta_{t+1}^{1/2}\sigma_t(\bx)
\right\},
\end{equation*}
where \(\beta_t>0\) is a tuning parameter that controls the amount of exploration and \(\sigma_t^2(\bx)=k_t(\bx,\bx)\) is the posterior variance.

In practice, the noise variance is often unknown, and the noise distribution need not be Gaussian. In such cases, the quantity \(\sigma^2\) in the posterior formulas can instead be viewed as a tuning parameter. 
From a frequentist perspective, this interpretation identifies the posterior mean $\mu_t(\bx)$ with the kernel ridge regression estimator, with \(\sigma^2\) playing the role of a regularization parameter \citep[Chapter~6]{RasmussenWilliams06}. 
Since this regularization parameter may depend on \(t\), the sample size used in the regression, we denote it by \(\rho_t\) to distinguish it from the true noise variance. Accordingly, when describing GP-UCB in Algorithm~\ref{algo:GP-UCB-eps}, we replace \(\sigma^2\) in \eqref{eq:posterior-mean} and \eqref{eq:posterior-cov} by \(\rho_t\).

\begin{algorithm}[ht]
  \caption{GP-UCB}\label{algo:GP-UCB-eps}
  \setlength{\abovedisplayskip}{1pt}
\setlength{\belowdisplayskip}{1pt}
  \SetAlgoLined
  \DontPrintSemicolon
  \SetKwInOut{Input}{Input}
  \Input{Kernel $k$ and tuning parameters $(\rho_t,\beta_t)$.}
  \For{all $t=0, 1, \ldots$}{
    Compute $\mu_t(\bx;\rho_t)$ and $\sigma_t^2(\bx;\rho_t)$ as
    \begin{align}
      \mu_t(\bx;\rho_t) = &\kb_t^\intercal(\bx)(\Kb_t+\rho_{t+1} \Ib_t)^{-1}\by_t, \label{eq:update-mu}\\
      \sigma_{t}^2(\bx;\rho_{t+1}) = & \bk(\bx, \bx) - \kb_t^\intercal(\bx)(\Kb_t+\rho_t \Ib_t)^{-1}\kb_t(\bx),\label{eq:update-sigma}
    \end{align}
    with $\mu_0(\bx;\rho_0) = 0$ and $\sigma_0^2(\bx;\rho_0) = k(\bx,\bx)$.\;
    Select
    $\bx_{t+1} \in \argmax\limits_{\bx\in \cD} \bigl\{\mu_{t}(\bx;\rho_{t+1}) + \beta_{t+1}^{1/2} \sigma_t(\bx;\rho_{t+1})\bigr\}$.\; 
    Observe $y_{t+1} = f(\bx_{t+1}) + \varepsilon_{t+1}$.\;
  }
\end{algorithm}

\subsection{Regret Analysis in the Frequentist Setting}

We analyze GP-UCB in the frequentist setting.
Rather than assuming that \(f\) is a sample path of a Gaussian process as in the Bayesian formulation, 
we assume that \(f\) belongs to the RKHS associated with the kernel $k$, defined as follows. 
We further assume that \(k\) is \emph{stationary}: that is, there exists a function \(\Psi:\cD\to\RR\) such that $k(\bx, \bx') = \Psi(\bx - \bx')$ for all $\bx, \bx'\in\cD$. 
Thus, \(k(\bx,\bx')\) depends on \(\bx\) and \(\bx'\) only through their difference. Throughout the paper, we will also refer to \(\Psi\) as the kernel when no confusion is likely to arise.

We denote by \(\mathcal{N}_{\Psi}(\cD)\) the RKHS associated with \(\Psi\). It consists of functions of the form
$g(\bx) = \sum_{j=1}^\infty c_j \Psi(\bx - \bx_j)$, where $c_j \in \RR$ and $\bx_j \in \cD$ for $j\geq 1$.
The space is equipped with the inner product $\langle \cdot, \cdot \rangle_{\mathcal{N}_{\Psi}(\cD)}$ defined as:
\[\langle g_1, g_2 \rangle_{\mathcal{N}_{\Psi}(\cD)} \coloneqq \sum_{j=1}^\infty\sum_{l=1}^\infty c_j c_l' \Psi(\bx_j -\bx_l'),\]
where $g_1(\bx) = \sum_{j=1}^\infty c_j \Psi(\bx - \bx_j)$ and $g_2(\bx) = \sum_{l=1}^\infty c_l' \Psi(\bx - \bx_l')$.
This inner product satisfies the \emph{reproducing property}: $\langle g, \Psi(\bx-\cdot)\rangle_{ \mathcal{N}_{\Psi}(\cD)} = g(\bx)$ for all $\bx\in \cD$.
It also induces the norm $\|\cdot\|_{\mathcal{N}_{\Psi}(\cD)}$ defined as $\|g\|_{\mathcal{N}_{\Psi}(\cD)} \coloneqq \sqrt{\langle g, g \rangle_{\mathcal{N}_{\Psi}(\cD)}} $. 
We refer to \cite{ScholkopfSmola02} for background to RKHS theory.

In this paper, we focus on the Mat\'ern class of kernels, given by
\begin{align}
\Psi(\bx-\bx')
\coloneqq
\frac{1}{\Gamma(\nu)2^{\nu-1}}
\biggl(\frac{2\sqrt{\nu}\|\bx-\bx'\|_2}{\ell}\biggr)^\nu
K_\nu\biggl(\frac{2\sqrt{\nu}\|\bx-\bx'\|_2}{\ell}\biggr),
\label{eq:matern}
\end{align}
where \(\nu>0\) is the \emph{smoothness} parameter, \(\ell>0\) is the length-scale parameter, \(\Gamma(\cdot)\) is the gamma function, \(K_\nu(\cdot)\) is the modified Bessel function of the second kind of order \(\nu\), and \(\|\cdot\|_2\) denotes the Euclidean norm. 


The regret performance of GP-UCB has been a central topic in the literature on Gaussian process bandits and Bayesian optimization. A substantial body of work has sought to sharpen regret upper bounds for GP-UCB and its variants in the frequentist setting with Mat\'ern kernels; see, for example, \cite{chowdhury2017kernelized}, \cite{JanzBurtGonzalez20}, \cite{VakiliKhezeliPicheny21}, and \cite{WhitehouseWuRamdas23}. These works have progressively improved the known upper bounds, from
\[
\sup_{\|f\|_{\mathcal{N}_{\Psi}(\cD)}\leq B} \EE[\cR(T;f,\mathrm{GP\mbox{-}UCB})] = 
\widetilde{O}\Bigl(T^{\frac{\nu + 3d(d+1)/2}{2\nu +d(d+1)}}\Bigr),
\]
obtained by \cite{srinivas2010gaussian} under the condition \(\nu>\frac{d(d+1)}{2}\), to
\begin{equation}\label{eq:best-upper-bounds}
  \sup_{\|f\|_{\mathcal{N}_{\Psi}(\cD)}\leq B} \EE[\cR(T;f,\mathrm{GP\mbox{-}UCB})]
  =
  \widetilde{O}\Bigl(T^{\frac{\nu+2d}{2\nu+2d}}\Bigr),
\end{equation}
proved by \cite{WhitehouseWuRamdas23} for \(\nu>\frac{1}{2}\). 
Here, $B$ is an arbitrary fixed constant and 
$\widetilde{O}(\cdot)$ denotes an asymptotic upper bound up to logarithmic factors. 
To our knowledge, \eqref{eq:best-upper-bounds} is the best regret upper bound currently available for GP-UCB.

Despite this progress, the regret optimality of GP-UCB remains unresolved. In particular, for the class of functions \(f\in \mathcal{N}_{\Psi}(\cD)\) with bounded RKHS norm, \cite{scarlett2017lower} established the minimax lower bound
\begin{equation}\label{eq:lower-bound}
  \inf_{\mathcal{A}} \sup_{\|f\|_{\mathcal{N}_{\Psi}(\cD)}\leq B}\EE[\cR(T;f,\mathcal{A})]
  =
  \Omega\Bigl(T^{\frac{\nu+d}{2\nu+d}}\Bigr),
\end{equation}
for optimization with noisy observations under i.i.d.\ Gaussian noise, where \(\Omega(\cdot)\) denotes an asymptotic lower bound. Comparing \eqref{eq:best-upper-bounds} and \eqref{eq:lower-bound} shows that a polynomial gap in \(T\) remains between the best known upper bound for GP-UCB and the minimax lower bound.
Although the steady improvement of upper bounds may suggest that GP-UCB is approaching minimax optimality, our results show that such an interpretation should be treated with caution by establishing a regret lower bound specific to GP-UCB itself.

\section{Main Result}\label{sec:mainresult}

Many regret upper-bound analyses of GP-UCB assume that the exploration and regularization sequences scale polynomially up to logarithmic factors, and rely on a uniform confidence bound to control the error in approximating \(f(\bx)\) by \(\mu_t(\bx;\rho_{t+1})\) \citep{srinivas2010gaussian,chowdhury2017kernelized,WhitehouseWuRamdas23}. Motivated by these works, we impose the following two assumptions.

\begin{assumption}\label{ass:seq}
    The sequences \(\{\beta_t\}_{t=1}^T\) and \(\{\rho_t\}_{t=1}^T\) are positive and nondecreasing, with \(\beta_t \ge 1\). There exist constants \(0 < c_\beta \le C_\beta < \infty\), \(0 < c_\rho \le C_\rho < \infty\), and exponents \(\theta_j \ge 0\), \(j=1,2,3,4\), all independent of \(T\), such that for all \(3 \le t \le T\),
\[
    c_\beta t^{\theta_1}(\log t)^{\theta_3}\le\beta_t\le C_\beta t^{\theta_1}(\log t)^{\theta_3},
\]
and
\[
    c_\rho t^{\theta_2}(\log t)^{\theta_4}\le\rho_t\le C_\rho t^{\theta_2}(\log t)^{\theta_4}.
\]
In addition, these exponents satisfy $\frac{4\nu+d}{2\nu}\theta_1 + \theta_2 < 1$.
\end{assumption}

The last condition in Assumption~\ref{ass:seq} is a technical requirement ensuring that a correction term in our regret lower bound remains nonzero (Section~\ref{sec:packing-size}); otherwise, the bound would become trivial. Nevertheless, this requirement is mild. In particular, for the choice of \(\beta_t\) and \(\rho_t\) considered by \citet{WhitehouseWuRamdas23}, it is satisfied for every \(\nu>0\).

\begin{assumption}[Uniform Confidence Bound]\label{ass:CI}
  There exist constants \(p_0 \in (0, 1)\) and \(\eta\in(0,1)\), independent of \(T\), such that the event $\cE_T$ holds with probability at least \(p_0\),
  where \[
    \cE_T:=\left\{|\mu_{t-1}(\bx; \rho_t)-f(\bx)|\le\eta\beta_t^{1/2}\sigma_{t-1}(\bx; \rho_t),\quad \forall \bx\in \cD,\ \forall t\le T\right\}.
  \]
\end{assumption}

As discussed in \cite{WhitehouseWuRamdas23}, Assumption~\ref{ass:CI} is essential to the regret analysis of GP-UCB. 

\begin{example}[\citet{srinivas2010gaussian} and \citet{chowdhury2017kernelized}]\label{example1} 
In these two studies, \(\rho_t\) is chosen to be constant, while \(\beta_t\) is taken to be of the same order as the so-called \emph{maximum information gain} \(\gamma_t\), which measures, in an information-theoretic sense, the largest possible reduction in uncertainty about \(f\) obtainable from \(t\) noisy observations. \citet{VakiliKhezeliPicheny21} showed that, when the regularization parameter \(\rho_t\) is constant, 
$\gamma_t=\widetilde{O}\bigl(t^{d/(2\nu+d)}\bigr)$,
so \(\beta_t\) grows polynomially in \(t\), up to logarithmic factors. 
Furthermore, Theorem~3 of \citet{srinivas2010gaussian} and Theorem~2 of \citet{chowdhury2017kernelized} established a uniform confidence bound of the form given in Assumption~\ref{ass:CI}.

\end{example}

\begin{example}[\citet{WhitehouseWuRamdas23}]\label{example2} 
This paper  refines earlier analyses by characterizing how the maximum information gain \(\gamma_t\) depends on the regularization parameter \(\rho_t\). For Mat\'ern kernels, the authors chose
$\rho_t=\widetilde{O}(t^{d/(2\nu+2d)})$,
and set \(\beta_t\) so that the effective optimism level \(L_t=\rho_t\beta_t\) satisfies
$L_t=\widetilde{O}(\gamma_t+\rho_t)$.
Since their bounds imply that \(\gamma_t\) also scales polynomially in \(t\) up to logarithmic factors, it follows that both \(\rho_t\) and \(\beta_t\) do as well. In addition, Lemma~4 of \citet{WhitehouseWuRamdas23} established a uniform confidence ellipsoid in RKHS norm; combined with the reproducing property and the Cauchy--Schwarz inequality, this yields a uniform confidence bound of the form given in Assumption~\ref{ass:CI}.


\end{example}

\begin{theorem}\label{thm:main}
    Let \(\Psi\) be the Mat\'ern kernel in \eqref{eq:matern}. Consider GP-UCB (Algorithm~\ref{algo:GP-UCB-eps}) with kernel $\Psi$ and tuning parameters \((\rho_t, \beta_t)\).  
    Suppose Assumptions~\ref{ass:seq} and~\ref{ass:CI} hold. Then, for all sufficiently large \(T\), there exists \(f\in\cN_{\Psi}(\cD)\), with \(\norm{f}_{\cN_{\Psi}(\cD)}\le B\), such that
    \[        \EE[\cR(T;f,\mathrm{GP\mbox{-}UCB})] = \Omega\!\left(T^{(\nu+d)/(2\nu+d)}L_T^{\nu/(2\nu+d)}\right).
    \]
\end{theorem}

\begin{proof}{Proof Sketch.} 
The proof of Theorem~\ref{thm:main} proceeds by constructing a family of localized hard instances (Section~\ref{sec:construction}) and then showing that GP-UCB cannot eliminate the resulting alternatives quickly enough without incurring substantial regret (Section~\ref{sec:analysis}). Specifically, we place many well-separated Mat\'ern bumps in the domain and designate one of them as the true bump, whose center is the true maximizer. The other bumps are called inactive bumps: they are not part of the true objective, but they represent competing alternatives that GP-UCB must rule out from the observations. The true core set is then defined as the small high-value neighborhood around the true maximizer. Hence, sampling inside this neighborhood can incur little regret, whereas sampling outside it incurs regret bounded below by a fixed fraction of the bump height. 
As long as an inactive bump
remains insufficiently sampled, the confidence event guarantees that its acquisition value stays competitive with that of the true bump. Hence, if GP-UCB continues to sample within the true core set 
around the maximizer while unresolved alternatives remain, then those core-set samples must occur at points with appreciably large posterior variance. Equivalently, the number of samples that can be taken inside the core set is limited. It follows that many samples must be placed outside the core set, thereby incurring substantial regret.
\halmos
\end{proof}

\smallskip

Theorem~\ref{thm:main} shows that if the effective optimism level \(L_T\) grows polynomially, then the regret lower bound is strictly larger than the minimax rate \(T^{(\nu+d)/(2\nu+d)}\). For example, \(L_T=\widetilde{O}(T^{d/(2\nu+d)})\) in Example~\ref{example1}, whereas \(L_T=\widetilde{O}(T^{d/(2\nu+2d)})\) in Example~\ref{example2}. After verifying the final condition in Assumption~\ref{ass:seq}, Theorem~\ref{thm:main} further implies that the choices of \((\rho_t,\beta_t)\) in Example~\ref{example1} cannot make GP-UCB minimax-optimal whenever \(\nu > \frac{1+\sqrt{5}}{4}d\), while those in Example~\ref{example2} cannot do so for any \(\nu>0\).

Consequently, for GP-UCB to achieve the minimax rate up to logarithmic factors, it is necessary that \(L_T\) grow at most polylogarithmically whenever one adopts the canonical regret-analysis framework initiated by \citet{srinivas2010gaussian} and used throughout much of the subsequent literature, namely, one based on establishing a uniform error bound of the form given in Assumption~\ref{ass:CI}; see \citet{VakiliScarlettJavidi21} for an overview of this approach.

\section{Construction of Hard Instances}\label{sec:construction}

\subsection{Localized Mat\'ern Bumps}\label{subsec:bumps}

We use the standard fact that the Mat\'ern RKHS on \(\cD\) is norm-equivalent to the Sobolev space \(H^{\nu+d/2}(\cD)\); see \citet[Corollary 10.48]{Wendland2004}. Let \(M\) be an integer to be chosen later. Choose points \(\bz_1,\ldots,\bz_M\in \cD\) with separation at least \(c h\), where
\[
  h\asymp M^{-1/d}.
\]
We choose all \(\bz_j\)'s in the interior of \(\cD\) satisfying the condition that balls of radius \(c h\) around the \(\bz_j\)'s are also contained in \(\cD\).

Let \(\psi\in C_c^\infty(\RR^d)\) be a strictly decreasing function and satisfy \(0\le\psi\le1\), \(\psi(0)=1\), and \(\operatorname{supp}(\psi)\subset B(0,1/4)\). Fix a small constant \(\varrho\in(0,1/8)\), and define
\[
  \chi_\varrho := 1-\sup_{\norm{u}\ge \varrho}\psi(u)>0.
\]
Define
\[
  \Delta=a_0h^\nu, 
\]
where \(a_0>0\) is a sufficiently small constant to be determined later, and define
\[
  g_j(\bx)=\Delta\psi\!\left(\frac{\bx-\bz_j}{h}\right),\qquad j=1,\ldots,M.
\]
Figure~\ref{fig:hard-instance} illustrates these functions in the case \(d=1\), and Lemma~\ref{lem:bump-scaling} characterizes their key properties.

\begin{figure}[ht]
  \centering
  \begin{subfigure}[t]{0.55\textwidth}
        \begin{tikzpicture}[x=1cm,y=1cm]
    \node[font=\small\bfseries, anchor=west] at (0.15,2.15) {$d=1$};

    \draw[thick] (0.4,0.9) -- (7.3,0.9);
    \draw[thick] (0.4,0.72) -- (0.4,1.08);
    \draw[thick] (7.3,0.72) -- (7.3,1.08);
    \node[anchor=south] at (3.85,2) {$\cD=[0,1]$};

    \filldraw[fill=blue!25, draw=blue!60!black, thick]
      (1.05,0.9) parabola bend (1.40,1.82) (1.75,0.9) -- cycle;
    \begin{scope}
      \clip (1.22,0.9) rectangle (1.58,1.9);
      \fill[blue!60!black]
        (1.05,0.9) parabola bend (1.40,1.82) (1.75,0.9) -- cycle;
    \end{scope}

    \fill[black] (1.40,0.9) circle (0.035);
    \fill[black] (2.70,0.9) circle (0.035);
    \fill[black] (3.85,0.9) circle (0.035);
    \fill[black] (5.30,0.9) circle (0.035);
    \fill[black] (6.50,0.9) circle (0.035);

    \node[blue!80!black, anchor=west] at (0.5,1.62) {$g_1$};
    \node[anchor=north] at (1.40,0.84) {$\bz_1$};
    \node[anchor=north] at (2.70,0.84) {$\bz_2$};
    \node[anchor=north] at (3.85,0.84) {$\bz_3$};
    \node[anchor=north] at (5.30,0.84) {$\bz_j$};
    \node[anchor=north] at (6.50,0.84) {$\bz_M$};
    \node[gray!70!black] at (4.58,1.18) {$\cdots$};

    \draw[<->, gray!70!black] (1.40,0.42) -- (2.70,0.42);
    \node[gray!70!black, fill=white, inner sep=1pt] at (2.05,0.58) {$\asymp h$};
  \end{tikzpicture}
  \end{subfigure}

    \begin{subfigure}[t]{0.55\textwidth}
        \begin{tikzpicture}[x=1cm,y=1cm]
    \node[font=\small\bfseries, anchor=west] at (0.15,2.15) {$d=1$};

    \draw[thick] (0.4,0.9) -- (7.3,0.9);
    \draw[thick] (0.4,0.72) -- (0.4,1.08);
    \draw[thick] (7.3,0.72) -- (7.3,1.08);


    \filldraw[fill=gray!25, draw=gray!60]
      (2.35,0.9) parabola bend (2.70,1.82) (3.05,0.9) -- cycle;

    \fill[black] (1.40,0.9) circle (0.035);
    \fill[black] (2.70,0.9) circle (0.035);
    \fill[black] (3.85,0.9) circle (0.035);
    \fill[black] (5.30,0.9) circle (0.035);
    \fill[black] (6.50,0.9) circle (0.035);

    \node[black, anchor=west] at (1.6,1.62) {$g_2$};
    \node[anchor=north] at (1.40,0.84) {$\bz_1$};
    \node[anchor=north] at (2.70,0.84) {$\bz_2$};
    \node[anchor=north] at (3.85,0.84) {$\bz_3$};
    \node[anchor=north] at (5.30,0.84) {$\bz_j$};
    \node[anchor=north] at (6.50,0.84) {$\bz_M$};
    \node[gray!70!black] at (4.58,1.18) {$\cdots$};

    \draw[<->, gray!70!black] (1.40,0.42) -- (2.70,0.42);
    \node[gray!70!black, fill=white, inner sep=1pt] at (2.05,0.58) {$\asymp h$};
  \end{tikzpicture}

  \end{subfigure}

      \begin{subfigure}[t]{0.55\textwidth}
        \begin{tikzpicture}[x=1cm,y=1cm]
    \node[font=\small\bfseries, anchor=west] at (0.15,2.15) {$d=1$};

    \draw[thick] (0.4,0.9) -- (7.3,0.9);
    \draw[thick] (0.4,0.72) -- (0.4,1.08);
    \draw[thick] (7.3,0.72) -- (7.3,1.08);


    \filldraw[fill=gray!25, draw=gray!60]
      (6.20,0.9) parabola bend (6.50,1.82) (6.80,0.9) -- cycle;

    \fill[black] (1.40,0.9) circle (0.035);
    \fill[black] (2.70,0.9) circle (0.035);
    \fill[black] (3.85,0.9) circle (0.035);
    \fill[black] (5.30,0.9) circle (0.035);
    \fill[black] (6.50,0.9) circle (0.035);

    \node[black, anchor=west] at (5.5,1.62) {$g_M$};
    \node[anchor=north] at (1.40,0.84) {$\bz_1$};
    \node[anchor=north] at (2.70,0.84) {$\bz_2$};
    \node[anchor=north] at (3.85,0.84) {$\bz_3$};
    \node[anchor=north] at (5.30,0.84) {$\bz_j$};
    \node[anchor=north] at (6.50,0.84) {$\bz_M$};
    \node[gray!70!black] at (4.58,1.18) {$\cdots$};

    \draw[<->, gray!70!black] (1.40,0.42) -- (2.70,0.42);
    \node[gray!70!black, fill=white, inner sep=1pt] at (2.05,0.58) {$\asymp h$};
  \end{tikzpicture}

  \end{subfigure}
  
  \caption{Schematic illustrations of the hard-instance construction with \(d=1\). The blue bump centered at \(\bz_1\) is the true maximizer, and the darker inner region denotes the true core \(C^\star=B(\bz_1,\varrho h)\). The gray bumps represent inactive alternatives centered at well-separated locations \(\bz_j\). Their supports are localized on scale \(h\), while pulls outside \(C^\star\) incur regret at least \(\chi_\varrho\Delta\).}
  \label{fig:hard-instance}
\end{figure}

\begin{lemma}\label{lem:bump-scaling}
  For \(a_0>0\) sufficiently small, \(\norm{g_j}_{\cN_{\Psi}(\cD)}\le B\) for every \(j\). Moreover, 
  \[
    g_j(\bz_j)=\Delta, \qquad \Delta\asymp M^{-\nu/d}.
  \]
\end{lemma}

\begin{proof}{Proof.}
  Since \(\cN_{\Psi}(\cD)\) is norm-equivalent to \(H^{\nu+d/2}(\cD)\), it suffices to control the Sobolev norm. Let
  \[
    \psi_{j,h}(\bx)=\psi\!\left(\frac{\bx-\bz_j}{h}\right).
  \]
  For \(s=\nu+d/2\), we have
  \[
    \norm{\psi_{j,h}}_{H^s(\cD)}\le C h^{d/2-s}=C h^{-\nu}.
  \]
  Therefore,
  \[
    \norm{g_j}_{\cN_{\Psi}(\cD)}\le C\Delta h^{-\nu} = Ca_0.
  \]
  Choosing \(a_0\le B/C\) gives \(\norm{g_j}_{\cN_{\Psi}(\cD)}\le B\). In addition, \(g_j(\bz_j)=\Delta\psi(0)=\Delta\). Since \(h\asymp M^{-1/d}\), we have \(\Delta=a_0h^\nu\asymp M^{-\nu/d}\). \halmos 
\end{proof}

We now fix the hard function
\[
  f=g_1.
\]
Then \(f(\bz_1)=\Delta\), and, for every \(j\ne1\), \(f(\bz_j)=0\). Define the true core ball
\[
  C^\star=B(\bz_1,\varrho h).
\]
By the definition of \(\chi_\varrho\), if \(\bx\notin C^\star\), then
\[
  f(\bx)\le (1-\chi_\varrho)\Delta.
\]
Consequently, every \(\bx_t\) outside \(C^\star\) incurs regret at least
\begin{align}\label{eq:out_reg}
  f(\bx^\star)-f(\bx_t) \ge \chi_\varrho\Delta.
\end{align}

\subsection{Posterior Variance for Unresolved Bumps}\label{subsec:variance}

For \(j\ne 1\), define the support of the \(j\)th inactive bump by
\[
  A_j:=\operatorname{supp}(g_j),
\]
and let
\[
  N_j(t):=\sum_{s=1}^t\one\{\bx_s\in A_j\}.
\]
The supports \(A_j\) are disjoint by construction.

We use the following variational formula for the posterior standard deviation.

\begin{lemma}\label{lem:variational}
  For any finite sample set \(S=\{\bx_1,\ldots,\bx_n\}\subset\cD\) and any \(\lambda>0\),
  \begin{align}\label{eq:lemvar}
    \sigma_{S,\lambda}(\bx) = \sup_{u\in\cN_{\Psi}(\cD)} \frac{|u(\bx)|}{\left(\norm{u}_{\cN_{\Psi}(\cD)}^2+ \lambda^{-1}\sum_{s=1}^n u(\bx_s)^2\right)^{1/2}}.
  \end{align}
\end{lemma}

\begin{proof}{Proof.}
  Let \(\cH:=\cN_{\Psi}(\cD)\). Equip \(\cH\) with the data-augmented inner product
  \[
    \inner{u}{v}_{S,\lambda} = \inner{u}{v}_{\cH} + \lambda^{-1}\sum_{s=1}^n u(\bx_s)v(\bx_s),
  \]
  with induced norm 
  \[
    \|u\|_{S,\lambda} = \left(\norm{u}_{\cH}^2+ \lambda^{-1}\sum_{s=1}^n u(\bx_s)^2\right)^{1/2}.
  \]
  Hence the right-hand side of \eqref{eq:lemvar} is the operator norm of the point-evaluation functional
  \[
    L_{\bx}:\cH\to\RR, \qquad L_{\bx}(u)=u(\bx)
  \]
  on the Hilbert space \((\cH,\inner{\cdot}{\cdot}_{S,\lambda})\). By the Riesz representation theorem, there exists a unique \(r_{\bx}\in\cH\) such that
  \begin{align}\label{eq:pfvar_Riesz}
    u(\bx)=\inner{u}{r_{\bx}}_{S,\lambda} \qquad \forall u\in\cH,
  \end{align}
  and moreover
  \[
    \sup_{u\in\cH}\frac{|u(\bx)|}{\|u\|_{S,\lambda}}=\|r_{\bx}\|_{S,\lambda}.
  \]

By the definition of the augmented inner product \(\inner{\cdot}{\cdot}_{S,\lambda}\), \eqref{eq:pfvar_Riesz} implies
 \begin{align}\label{eq:pfvar_Riesz2}
    u(\bx)=\inner{u}{r_{\bx}}_{\cH}+\lambda^{-1}\sum_{s=1}^n u(\bx_s)r_{\bx}(\bx_s).
\end{align}
  Using the reproducing property \(u(\bz)=\inner{u}{\Psi(\bz-\cdot)}_{\cH}\), \eqref{eq:pfvar_Riesz2} further implies that
  \begin{align}\label{eq:pfvar_Riesz3}
    \inner{u}{\Psi(\bx-\cdot)}_{\cH} = & \inner{u}{r_{\bx}}_{\cH}+\lambda^{-1}\sum_{s=1}^n \inner{u}{\Psi(\bx_s-\cdot)}_{\cH}r_{\bx}(\bx_s)\nonumber\\
    =  & \inner{u}{r_{\bx}+\lambda^{-1}\sum_{s=1}^n r_{\bx}(\bx_s)\Psi(\bx_s-\cdot)}_{\cH} \qquad \forall u\in\cH.
  \end{align}
Since \eqref{eq:pfvar_Riesz3} holds for all $u\in \cH$, we must have
  \[
    \Psi(\bx-\cdot) = r_{\bx}+\lambda^{-1}\sum_{s=1}^n r_{\bx}(\bx_s)\Psi(\bx_s-\cdot).
  \]
  Equivalently, \(r_{\bx}\) has the form
  \[
    r_{\bx}=\Psi(\bx-\cdot)-\sum_{s=1}^n a_s\Psi(\bx_s-\cdot),
  \]
  where \(a_s=\lambda^{-1}r_{\bx}(\bx_s)\). Evaluating this identity at the sample points \(\bx_1,\ldots,\bx_n\) gives the linear system
  \[
    (K_S+\lambda I)a=\kb_S(\bx),
  \]
  where \(a=(a_1,\ldots,a_n)^\intercal, \kb_S(\bx)=(\Psi(\bx-\bx_1),\ldots,\Psi(\bx-\bx_n))^\intercal, K_S=(\Psi(\bx_i-\bx_j))_{1\le i,j\le n}
  \). Hence
  \[
    a=(K_S+\lambda I)^{-1}\kb_S(\bx).
  \]
  Taking \(u=r_{\bx}\) in the identity \(u(\bx)=\inner{u}{r_{\bx}}_{S,\lambda}\) yields
  \[
    \|r_{\bx}\|_{S,\lambda}^2=r_{\bx}(\bx).
  \]
  Substituting the expression for \(r_{\bx}\) and the formula for \(a\), we obtain
  \[
    \|r_{\bx}\|_{S,\lambda}^2 = \Psi(\mathbf 0)-\kb_S^\intercal(\bx)(K_S+\lambda I)^{-1}\kb_S(\bx).
  \]
  This is exactly \(\sigma_{S,\lambda}^2(\bx)\). Taking square roots proves the result. \halmos 
\end{proof}

Let \(\cI_T:=\{\lceil T/2\rceil,\ldots,T\}\). Next, we show a fact implied by Assumption~\ref{ass:seq} that will be used throughout the proof.

\begin{lemma}\label{lem:polydecay-seq}
  Under Assumption~\ref{ass:seq}, there exists a constant \(c_\star\in(0,1]\), independent of \(T\), such that for every \(t\in\cI_T\),
  \[
    \beta_t\ge c_\star\beta_T,\quad \rho_t\ge c_\star\rho_T, \quad L_t\ge c_\star L_T.
  \]
  Moreover, for all sufficiently large \(T\), the effective optimism level satisfies \(1\lesssim L_T\le T\). Let
  \[
    M_T:=\left(\frac{T}{L_T}\right)^{d/(2\nu+d)}, \qquad
    \ell_T:=\log\!\left(1+\frac{T}{\rho_T}\right),
  \]
  and
  \[
    \zeta_T:=\left(1+\beta_T^{1/(2\nu)}+\log(2+M_T)\right)^d.
  \]
  Then \(\zeta_T\ell_T/M_T\to0\).
\end{lemma}

\begin{proof}{Proof.}
  For \(t\in\cI_T\), \(t/T\ge1/2\), and \(\log t/\log T\) is bounded below by a positive universal constant. This gives \(\beta_t\ge c_\star\beta_T\) and \(\rho_t\ge c_\star\rho_T\), where \(c_\star>0\) depends only on \(c_\beta,C_\beta,c_\rho,C_\rho\) and \(\theta_j\)'s. Then \(L_t\ge c_\star L_T\) directly follows. Assumption~\ref{ass:seq}  gives \(L_T\asymp T^{\theta_1+\theta_2}(\log T)^{\theta_3+\theta_4}\). Since all \(\theta_j\)'s are nonnegative, it can be checked that \(\theta_1+\theta_2<1\), and thus \(1\lesssim L_T\le T\) for all sufficiently large \(T\). 

  Direct computation shows that \(M_T\asymp T^{d(1-\theta_1-\theta_2)/(2\nu+d)}(\log T)^{-d(\theta_3+\theta_4)/(2\nu+d)}\), \(\ell_T=O(\log T)\), and \(\zeta_T=O\!\left(T^{d\theta_1/(2\nu)}(\log T)^{C}\right)\) for some constant \(C\) depending only on \(d,\nu\), and \(\theta_j\)'s. By Assumption~\ref{ass:seq}, it can be checked that the polynomial exponent of \(\zeta_T\ell_T/M_T\) strictly negative, so \(\zeta_T\ell_T/M_T\to0\). \halmos
\end{proof}

The next lemma says that a bad bump remains optimistic during the terminal half of the horizon until it has been sampled on the order of \(L_T/\Delta^2\) times.

\begin{lemma}\label{lem:unresolved}
  There exist constants \(c_m>0\) and \(\kappa>0\) such that the following holds. Define
  \[
    m_\Delta := \left\lfloor c_m\frac{L_T}{\Delta^2}\right\rfloor.
  \]
  If \(t\in\cI_T\), \(j\ne1\), and \(N_j(t-1)\le m_\Delta\), then
  \[
    \beta_t^{1/2}\sigma_{t-1}(\bz_j; \rho_t) \ge \frac{1+\kappa}{1-\eta}\Delta.
  \]
  Consequently, on \(\cE_T\), the GP-UCB acquisition value at round \(t\) satisfies
  \[
    \mu_{t-1}(\bz_j; \rho_t)+\beta_t^{1/2}\sigma_{t-1}(\bz_j; \rho_t) \ge (1+\kappa)\Delta.
  \]
\end{lemma}

\begin{proof}{Proof.}
  Apply Lemma~\ref{lem:variational} with \(u=g_j\), \(\bx=\bz_j\), \(S=\{\bx_1,\ldots,\bx_{t-1}\}\), and \(\lambda=\rho_t\). Since \(g_j\) is supported on \(A_j\) and \(0\le g_j\le\Delta\),
  \[
    \sigma_{t-1}(\bz_j; \rho_t) \ge \frac{\Delta} {\left(\norm{g_j}_{\cN_{\Psi}(\cD)}^2+ \rho_t^{-1}N_j(t-1)\Delta^2\right)^{1/2}}.
  \]
  By Lemma~\ref{lem:bump-scaling}, \(\norm{g_j}_{\cN_{\Psi}(\cD)}\le C a_0\). If \(N_j(t-1)\le m_\Delta\), then Lemma~\ref{lem:polydecay-seq} gives
  \[
    \rho_t^{-1}N_j(t-1)\Delta^2 \le c_m\frac{L_T}{\rho_t} \le C c_m\beta_t.
  \]
  Therefore, we can choose \(a_0\) and then \(c_m\) small enough so that for some \(\kappa>0\),
  \[
    \beta_t^{1/2}\sigma_{t-1}(\bz_j; \rho_t) \ge \frac{\Delta}{\left(C^2a_0^2/\beta_t+C c_m\right)^{1/2}} \ge \frac{\Delta}{\left(C^2a_0^2+C c_m\right)^{1/2}}\ge \frac{1+\kappa}{1-\eta}\Delta,
  \]
 Under the event \(\cE_T\), since \(f(\bz_j)=0\), Assumption~\ref{ass:CI} implies that
  \[
    \mu_{t-1}(\bz_j; \rho_t) \ge -\eta\beta_t^{1/2}\sigma_{t-1}(\bz_j; \rho_t).
  \]
  Thus
  \[
    \mu_{t-1}(\bz_j; \rho_t)+\beta_t^{1/2}\sigma_{t-1}(\bz_j; \rho_t) \ge (1-\eta)\beta_t^{1/2}\sigma_{t-1}(\bz_j; \rho_t)\ge (1+\kappa)\Delta,
  \]
  which finishes the proof.  \halmos 
\end{proof}

\subsection{A Local Log-Determinant Bound}\label{subsec:local-logdet}

The proof also requires a bound on the number of high-posterior-variance pulls that can occur in a small ball. We use a version with an explicit approximation parameter.

\begin{lemma}\label{lem:upperDet}
  Let $A,B\in\RR^{n\times n}$ be two symmetric, positive semidefinite matrices. Thus,
  \begin{align*}
    \det(I_n+A+B)\le \det(I_n+A)\det(I_n+B).
  \end{align*}
\end{lemma}
\begin{proof}{Proof.}
  Let $\alpha_1\ge \alpha_2 \ge ... \ge\alpha_n>0$ and $\beta_1\ge \beta_2 \ge ... \ge\beta_n>0$ be eigenvalues of matrices $A$ and $B$, respectively. Therefore, we have
  \begin{align*}
    \det(I_n+A)\det(I_n+B) & = \prod_{i=1}^n(1+\alpha_i)(1+\beta_{n+1-i})\\
    & = \prod_{i=1}^n(1+\alpha_i+\beta_{n+1-i}+\beta_{n+1-i}\alpha_i)\\
    & \ge \prod_{i=1}^n(1+\alpha_i+\beta_{n+1-i})\\
    & \ge \det(I_n+A+B),
  \end{align*}
  where the last inequality is true because of Fiedler bound \citep{fiedler1971bounds}. \halmos 
\end{proof}

\begin{lemma}\label{lem:local-logdet}
  Let \(S_n=\{\bx_1,\ldots,\bx_n\}\subset B(\bz,r)\subset \cD\), and let \(K_{S_n}\) be the Mat\'ern kernel matrix on \(S_n\). For any \(A\ge1\) and any \(\lambda>0\), there exist constants \(C_1,C_2>0\), depending only on \(d\), \(\nu\), and the kernel, such that
  \[
    \log\det(I_n+\lambda^{-1}K_{S_n}) \le \zeta(A,r)\log(1+n/\lambda) + C_1\frac{n r^{2\nu}}{A^2\lambda},
  \]
  where
  \[
    \zeta(A,r) \le C_2\left(1+A^{1/\nu}+|\log r|\right)^d .
  \]
  In particular, for \(r=\varrho h\) and \(h^{2\nu}=\Delta^2/a_0^2\),
  \[
    \log\det(I_n+\lambda^{-1}K_{S_n}) \le \zeta(A,\varrho h)\log(1+n/\lambda) + C_1\varrho^{2\nu}\frac{n\Delta^2}{A^2\lambda}.
  \]
\end{lemma}

\begin{proof}{Proof.}
  A stationary Mat\'ern kernel has a spectral representation
  \[
    \Psi(\bx-\by)=\int_{\RR^d}e^{i\inner{\bx-\by}{\omega}}q(\omega)\,d\omega,
  \]
  where \(q(\omega)\asymp (1+\norm{\omega}^2)^{-(\nu+d/2)}\). Thus \(\Psi(\bx-\by)=\inner{\Phi(\bx)}{\Phi(\by)}_{L^2}\), with feature map
  \[
    \Phi(\bx)(\omega)=e^{i\inner{\bx}{\omega}}q(\omega)^{1/2}.
  \]
  Write \(\bx=z+u\), where \(\norm{u}\le r\). Fix \(A\ge1\) and set
  \[
    R=A^{1/\nu}r^{-1}.
  \]
  Hence,
  \begin{align}\label{eq:pf_lemdetub_hf}
    \int_{\norm{\omega}>R}q(\omega)\,d\omega \le C R^{-2\nu} = C\frac{r^{2\nu}}{A^2}.
  \end{align}

  For \(\norm{\omega}\le R\), we approximate \(e^{i\inner{u}{\omega}}\) by its Taylor expansion in \(\inner{u}{\omega}\). If \(\norm{u}\le r\), then the Cauchy-Schwarz inequality implies that
  \[
    |\inner{u}{\omega}|\le \norm{u}\,\norm{\omega}\le rR=A^{1/\nu}.
  \]
  Hence the remainder of Taylor expansion is bounded uniformly over \(\norm{u}\le r\) and \(\norm{\omega}\le R\) by
  \[
    \left|e^{i\inner{u}{\omega}}-\sum_{j=0}^m \frac{(i\inner{u}{\omega})^j}{j!}\right| \le \frac{e^{A^{1/\nu}}A^{(m+1)/\nu}}{(m+1)!}.
  \]
  Using Stirling's formula, we choose
  \[
    m\asymp 1+A^{1/\nu}+|\log r|,
  \]
  and thus the squared \(L^2(q)\) remainder is at most \(C r^{2\nu}/A^2\), uniformly for \(\norm{u}\le r\), i.e.,
  \begin{align}\label{eq:pf_lemdetub_lf}
    \int_{\norm{\omega}\le R}\left|e^{i\inner{u}{\omega}}-\sum_{j=0}^m \frac{(i\inner{u}{\omega})^j}{j!}\right|^2q(\omega)\,d\omega \le C\frac{r^{2\nu}}{A^2}.
  \end{align}
  Combining \eqref{eq:pf_lemdetub_hf} and \eqref{eq:pf_lemdetub_lf}, we can conclude that there exists a subspace \(V\subset L^2\) of dimension
  \[
    \dim(V) \le C\left(1+A^{1/\nu}+|\log r|\right)^d =:\zeta(A,r)
  \]
  such that
  \[
    \inf_{v\in V}\norm{\Phi(\bx)-v}_{L^2}^2 \le C\frac{r^{2\nu}}{A^2} \qquad \forall \bx\in B(z,r).
  \]

  Let \(P\) be the orthogonal projection in \(L^2\) onto \(V\), and decompose the kernel matrix as
  \[
    K_{S_n}=A_n+R_n,
  \]
  where
  \[
    (A_n)_{ij}=\inner{P\Phi(\bx_i)}{P\Phi(\bx_j)}_{L^2}, (R_n)_{ij}=\inner{(I-P)\Phi(\bx_i)}{(I-P)\Phi(\bx_j)}_{L^2}.
  \]
  Both \(A_n\) and \(R_n\) are positive semidefinite. Moreover, \(\rank(A_n)\le \zeta(A,r)\), \(\tr(A_n)\le Cn\), and
  \[
    \tr(R_n) = \sum_{i=1}^n\norm{(I-P)\Phi(\bx_i)}_{L^2}^2 \le C\frac{n r^{2\nu}}{A^2}.
  \]
  Lemma \ref{lem:upperDet} implies that
  \begin{align}\label{eq:pf_lemdetub_det1}
    \log\det(I_n+\lambda^{-1}K_{S_n}) \le \log\det(I_n+\lambda^{-1}A_n) + \log\det(I_n+\lambda^{-1}R_n).
  \end{align}
  Since \(A_n\) has rank at most \(\zeta(A,r)\) and trace at most \(Cn\),
    \begin{align}\label{eq:pf_lemdetub_det11}
    \log\det(I_n+\lambda^{-1}A_n) \le \zeta(A,r)\log(1+Cn/\lambda).
  \end{align}
  For the residual part,
    \begin{align}\label{eq:pf_lemdetub_det12}
    \log\det(I_n+\lambda^{-1}R_n) \le \lambda^{-1}\tr(R_n) \le C\frac{n r^{2\nu}}{A^2\lambda}.
  \end{align}
  Combining \eqref{eq:pf_lemdetub_det1}-\eqref{eq:pf_lemdetub_det12} gives the first claim. The second claim follows from \(r=\varrho h\) and \(h^{2\nu}=\Delta^2/a_0^2\), after absorbing \(a_0^{-2}\) into the constant.  \halmos 
\end{proof}

\section{Regret Analysis}\label{sec:analysis}

\subsection{Bounding the Number of Terminal Core Pulls}\label{subsec:core-pulls}

Let
\[
  U_t(\bx)=\mu_{t-1}(\bx; \rho_t)+\beta_t^{1/2}\sigma_{t-1}(\bx; \rho_t)
\]
be the GP-UCB acquisition function at the beginning of round \(t\). We first show that whenever an unresolved bad bump exists during the terminal half of the horizon, every pull in the true core must have large posterior variance.

\begin{lemma}\label{lem:core-high-var}
  Under the event \(\cE_T\), suppose \(t\in\cI_T\) and that at the beginning of round \(t\) there exists \(j\ne1\) such that \(N_j(t-1)\le m_\Delta\). If GP-UCB selects \(\bx_t\in C^\star\), then
  \[
    \sigma_{t-1}(\bx_t; \rho_t) \ge c_\sigma\frac{\Delta}{\beta_t^{1/2}}
  \]
  for a constant \(c_\sigma>0\).
\end{lemma}

\begin{proof}{Proof.}
  By Lemma~\ref{lem:unresolved},
  \[
    U_t(z_j) \ge (1+\kappa)\Delta.
  \]
  Because GP-UCB chooses a maximizer of \(U_t\), if \(\bx_t\in C^\star\), then
  \begin{align}\label{eq:pf_lem_corehv_eq1}
    U_t(\bx_t) \ge U_t(z_j) \ge (1+\kappa)\Delta.
  \end{align}
  Under the event \(\cE_T\), we have
  \[
    \mu_{t-1}(\bx_t; \rho_t) \le f(\bx_t)+\eta\beta_t^{1/2}\sigma_{t-1}(\bx_t; \rho_t).
  \]
  which, together with \(f(\bx_t)\le\Delta\), leads to
  \begin{align}\label{eq:pf_lem_corehv_eq2}
    U_t(\bx_t) = \mu_{t-1}(\bx_t; \rho_t)+\beta_t^{1/2}\sigma_{t-1}(\bx_t; \rho_t) \le \Delta+(1+\eta)\beta_t^{1/2}\sigma_{t-1}(\bx_t; \rho_t).
  \end{align}
  Combining \eqref{eq:pf_lem_corehv_eq1} and \eqref{eq:pf_lem_corehv_eq2} gives
  \[
    \beta_t^{1/2}\sigma_{t-1}(\bx_t; \rho_t) \ge \frac{\kappa}{1+\eta}\Delta.
  \]
  This finishes the proof by setting \(c_\sigma=\kappa/(1+\eta)\).  \halmos 
\end{proof}

We next use the local log-determinant bound to show that only selecting few such high-variance terminal core points are possible.

\begin{lemma}\label{lem:number-core}
  Let
  \[
    A_T=c_A(1+\beta_T^{1/2}), \qquad \ell_T=\log\!\left(1+\frac{T}{\rho_T}\right),
  \]
  where \(c_A\) is a sufficiently large constant. Define
  \[
    \zeta_T(h) := \left(1+A_T^{1/\nu}+|\log h|\right)^d .
  \]
  Under the event \(\cE_T\), the number of rounds \(t\in\cI_T\) at which GP-UCB select points in \(C^\star\) while at least one inactive bump satisfies \(N_j(t-1)\le m_\Delta\) is at most
  \[
    C\frac{L_T}{\Delta^2}\zeta_T(h)\ell_T.
  \]
\end{lemma}

\begin{proof}{Proof.}
  Consider the subsequence of rounds \(\tau_1<\cdots<\tau_n\) in \(\cI_T\) at which GP-UCB selects a point in \(C^\star\) and at least one inactive bump is unresolved in the sense that \(N_j(\tau_i-1)\le m_\Delta\). Let \(u_i=\bx_{\tau_i}\). By Lemma~\ref{lem:core-high-var}, we must have
  \[
    \sigma_{\tau_i-1}(u_i; \rho_{\tau_i}) \ge c_\sigma\frac{\Delta}{\beta_{\tau_i}^{1/2}} \ge c\frac{\Delta}{\beta_T^{1/2}},
  \]
  where the last step uses \(\beta_{\tau_i}\le\beta_T\). Let \(\widetilde\sigma_{i-1}(u_i; \rho_T)\) denote the posterior standard deviation at \(u_i\) computed using only the earlier selected core points \(u_1,\ldots,u_{i-1}\), with regularization \(\rho_T\). Adding observations can only decrease posterior variance, and increasing the regularization level can only increase posterior variance. Therefore
  \begin{align}\label{eq:pullub_sigmalb_1}
    \widetilde\sigma_{i-1}(u_i; \rho_T) \ge \sigma_{\tau_i-1}(u_i; \rho_T) \ge
    \sigma_{\tau_i-1}(u_i; \rho_{\tau_i}) \ge c\frac{\Delta}{\beta_T^{1/2}}.
  \end{align}
  The determinant identity for sequential Gaussian posterior variances gives
  \begin{align}\label{eq:pullub_deteq}
    \log\det(I_n+\rho_T^{-1}K_{u_{1:n}}) = \sum_{i=1}^n \log\left(1+\rho_T^{-1}\widetilde\sigma_{i-1}^2(u_i; \rho_T)\right).
  \end{align}
  This is because, by writing \(K_i:=K_{u_{1:i}}\), the block determinant formula gives
  \[
    \det(I_i+\rho_T^{-1}K_i)=\det(I_{i-1}+\rho_T^{-1}K_{u_1:i-1})\left(1+\rho_T^{-1}\widetilde\sigma_{i-1}^2(u_i; \rho_T)\right),
  \]
  and iterating over \(i=1,\ldots,n\) then taking logarithms yields \eqref{eq:pullub_deteq}. Therefore, \eqref{eq:pullub_sigmalb_1} implies that
  \begin{align}\label{eq:pullub_deteq_lb}
    \log\det(I_n+\rho_T^{-1}K_{u_{1:n}}) \ge n\log\left(1+c\frac{\Delta^2}{\rho_T\beta_T}\right) \ge c n\frac{\Delta^2}{L_T},
  \end{align}
  where the last inequality is because \(L_T=\rho_T\beta_T\), Lemma~\ref{lem:polydecay-seq} gives \(1\lesssim L_T\lesssim T\) and the construction below ensure \(\Delta^2/L_T\le C\). 

  On the other hand, all \(u_i\)'s belong to \(C^\star=B(\bz_1,\varrho h)\). Applying Lemma~\ref{lem:local-logdet} with \(A=A_T\) and \(\lambda=\rho_T\) yields
  \begin{align}\label{eq:pullub_deteq_ub1}
    \log\det(I_n+\rho_T^{-1}K_{u_{1:n}}) \le & \zeta(A_T,\varrho h)\log(1+n/\rho_T) + C\varrho^{2\nu}\frac{n\Delta^2}{A_T^2\rho_T}\nonumber\\
    \le & \zeta(A_T,\varrho h)\log(1+n/\rho_T) + C\varrho^{2\nu}\frac{\beta_T}{A_T^2}\frac{n\Delta^2}{L_T}\nonumber\\
    \le & \zeta(A_T,\varrho h)\log(1+n/\rho_T) + \frac{1}{2}c n\frac{\Delta^2}{L_T},
  \end{align}
  where in the last inequality, because \(A_T=c_A(1+\beta_T^{1/2})\), we can choose \(c_A\) large enough such that 
  \[
    C\varrho^{2\nu}\frac{\beta_T}{A_T^2}\frac{n\Delta^2}{L_T} \leq \frac{1}{2}c n\frac{\Delta^2}{L_T}.
  \]
  Combining \eqref{eq:pullub_deteq_lb} and \eqref{eq:pullub_deteq_ub1} gives
  \[
    \frac{1}{2}c n\frac{\Delta^2}{L_T} \le \zeta(A_T,\varrho h)\log(1+n/\rho_T).
  \]
  Since \(n\le T\), \(\log(1+n/\rho_T)\le \ell_T\). Also, because \(\varrho\) is fixed,
  \[
    \zeta(A_T,\varrho h) \le C\left(1+A_T^{1/\nu}+|\log h|\right)^d = C\zeta_T(h).
  \]
  Therefore
  \[
    n \le C\frac{L_T}{\Delta^2}\zeta_T(h)\ell_T,
  \]
  which finishes the proof. \halmos 
\end{proof}

\subsection{Choice of the Packing Size and Proof of Theorem~\ref{thm:main}}\label{subsec:choice-proof} \label{sec:packing-size}

We now choose the number of bumps. Set
\[
  M = \left\lfloor c_M\left(\frac{T}{L_T}\right)^{d/(2\nu+d)}\right\rfloor,
\]
where \(c_M>0\) is a sufficiently large constant. Then
\[
  \Delta=a_0M^{-\nu/d}\asymp \left(\frac{L_T}{T}\right)^{\nu/(2\nu+d)}, \quad \frac{L_T}{\Delta^2} \asymp T^{2\nu/(2\nu+d)}L_T^{d/(2\nu+d)} \asymp \frac{T}{M}.
\]
Let \(N_{\mathrm{out}}(T)\) be the number of points in \(\{\bx_t\}_{t=1}^T\) outside the true core \(C^\star\). Each such point incurs regret at least \(\chi_\varrho\Delta\), shown in \eqref{eq:out_reg}. We show that, under \(\cE_T\),
\[
  N_{\mathrm{out}}(T)\ge cT\left(1-C\frac{\zeta_T\ell_T}{M_T}\right)_+,
\]
where \(M_T=(T/L_T)^{d/(2\nu+d)}\), \(\ell_T=\log(1+T/\rho_T)\), and
\[
  \zeta_T = \left(1+\beta_T^{1/(2\nu)}+\log(2+M_T)\right)^d .
\]

There are two cases. 

\textit{Case 1.} Suppose that, by time \(T\), all inactive bumps are resolved at the terminal threshold, meaning that \(N_j(T)>m_\Delta\) for every \(j\ne1\). Then
\[
  N_{\mathrm{out}}(T) \ge \sum_{j\ne1}N_j(T) \ge (M-1)m_\Delta.
\]
By the definitions of \(M\) and \(m_\Delta\), and by choosing \(c_M\) large enough after \(c_m\) is fixed,
\[
  (M-1)m_\Delta \ge cT.
\]
Thus the desired lower bound on \(N_{\mathrm{out}}(T)\) holds in this case.

\textit{Case 2.} Suppose that at least one inactive bump remains unresolved at time \(T\); that is, there exists \(j\ne1\) such that \(N_j(T)\le m_\Delta\). Then for every \(t\in\cI_T\), the same bump satisfies \(N_j(t-1)\le m_\Delta\). Hence every selected point in \(C^\star\) is in the situation covered by Lemma~\ref{lem:number-core}. The number of selected points in \(C^\star\) during \(\cI_T\) is therefore at most
\[
  C\frac{L_T}{\Delta^2}\zeta_T(h)\ell_T \le C\frac{T}{M}\zeta_T\ell_T.
\]
Since \(|\cI_T|\ge T/2\), the number of selected points outside \(C^\star\), and hence \(N_{\mathrm{out}}(T)\), is at least
\[
  \frac{T}{2}-C\frac{T}{M}\zeta_T\ell_T \ge cT\left(1-C\frac{\zeta_T\ell_T}{M_T}\right)_+,
\]
after adjusting constants and using \(M\asymp M_T\).

Combining these two cases gives
\[
  \cR(T;f,\mathrm{GP\mbox{-}UCB}) \ge \chi_\varrho\Delta N_{\mathrm{out}}(T) \ge cT\Delta \left(1-C\frac{\zeta_T\ell_T}{M_T}\right)_+
\]
under the event \(\cE_T\). Since Assumption~\ref{ass:CI} gives \(\PP_f(\cE_T)\ge p_0\), for sufficiently large \(T\), the regret can be lower bounded by
\begin{align*}
\EE[\cR(T;f,\mathrm{GP\mbox{-}UCB})] \ge{}& \PP_f(\cE_T)\inf_{\cE_T}\cR(T;f,\mathrm{GP\mbox{-}UCB}) \\
\ge{}& cT\Delta \left(1-C\frac{\zeta_T\ell_T}{M_T}\right)_+ \ge \frac c2 T\Delta,    
\end{align*}
where the last inequality is because by Lemma~\ref{lem:polydecay-seq}, \(\zeta_T\ell_T/M_T\to0\). Hence, for sufficiently large \(T\), 
This proves Theorem~\ref{thm:main} by noting that
\[
  T\Delta \asymp T\left(\frac{L_T}{T}\right)^{\nu/(2\nu+d)} = T^{(\nu+d)/(2\nu+d)}L_T^{\nu/(2\nu+d)}.
\]

\section{Conclusion}\label{sec:conclusion}

We introduced the effective optimism level $L_t=\rho_t\beta_t$ as the quantity capturing how exploration and regularization jointly shape the behavior of GP-UCB. 
From this viewpoint, we proved a new regret lower bound for GP-UCB with Mat\'ern kernels under a uniform confidence assumption.
The main implication is clear: if $L_t$ grows polynomially in $t$, up to logarithmic factors, then GP-UCB cannot achieve the minimax-optimal regret rate.
Since this is the regime covered by most existing analyses, our result points to a real obstacle to proving minimax optimality for standard GP-UCB. In particular, the gap between current upper bounds and minimax lower bounds should not be viewed only as a limitation of existing proof techniques. It may also reflect a limitation of the algorithm itself. 
More broadly, our results suggest that any route to minimax-optimal guarantees for GP-UCB over Mat\'ern RKHS classes must either drive the effective optimism level much lower, possibly to a polylogarithmic scale, or depart from the standard GP-UCB formulation.

%
%
%




\bibliographystyle{informs2014} 
\bibliography{ref} 






\end{document}